\documentclass{sig-alternate}
\pdfoutput=1 %for posting to arXiv
\usepackage{blindtext, graphicx}
\usepackage{epsfig}
\usepackage{syntonly}
\usepackage{rotating}
\usepackage{amsmath}
\usepackage{setspace}	
\usepackage{verbatim} 
\usepackage{amssymb}
\usepackage{amsmath}
\usepackage{graphicx}
\usepackage{multirow}
\usepackage{colortbl}
\usepackage{enumerate}
\usepackage{color}
\usepackage{wrapfig}

\usepackage{algorithm2e}
\usepackage{cleveref}
\usepackage{comment}
\usepackage{ifthen}
\usepackage[dotinlabels]{titletoc}
\usepackage{colortbl}
\usepackage{array}
\usepackage{alltt}
\usepackage{bm}

\usepackage[normalem]{ulem}

\pdfoutput=1 %for posting to arXiv
\usepackage{blindtext, graphicx}
\usepackage{epsfig}
\usepackage{syntonly}
\usepackage{rotating}
\usepackage{amsmath}
\usepackage{setspace}
\usepackage{verbatim} 
\usepackage{amssymb}
\usepackage{amsmath}
\usepackage{graphicx}
\usepackage{multirow}
\usepackage{colortbl}
\usepackage{enumerate}
\usepackage{color}
\usepackage{wrapfig}
\usepackage{algorithm2e}
\usepackage{cleveref}
\usepackage{comment}
\usepackage{ifthen}
\usepackage[dotinlabels]{titletoc}
\usepackage{colortbl}
\usepackage{array}

\usepackage{alltt}
\usepackage{bm}
\usepackage{longtable,booktabs}
\usepackage{url}
\usepackage{courier}
\usepackage{cite}

\usepackage{tabularx}
\usepackage{footnote}

\usepackage{colortbl}
\usepackage{threeparttable}
\usepackage[free-standing-units]{siunitx}

\begin{document}

\CopyrightYear{2017} 
\setcopyright{acmlicensed}
\conferenceinfo{FPGA '17,}{February 22 - 24, 2017, Monterey, CA, USA}
\isbn{978-1-4503-4354-1/17/02}\acmPrice{\$15.00}
\doi{http://dx.doi.org/	10.1145/3020078.3021745}

\title{ESE: Efficient Speech Recognition Engine \\ with Sparse LSTM on FPGA}
\author{Song Han$^{1,2}$, Junlong Kang$^{2}$, Huizi Mao$^{1,2}$, Yiming Hu$^{2,3}$, Xin Li$^{2}$, Yubin Li$^{2}$, Dongliang Xie$^{2}$\\
Hong Luo$^{2}$, Song Yao$^{2}$, Yu Wang$^{2,3}$, Huazhong Yang$^{3}$ and William J. Dally$^{1,4}$\\
$^{1}$ Stanford University, $^{2}$ DeePhi Tech, $^{3}$ Tsinghua University, $^{4}$ NVIDIA\\
\emph{$^{1}$ \{songhan,dally\}@stanford.edu, $^{2}$ song.yao@deephi.tech, $^{3}$ yu-wang@mail.tsinghua.edu.cn}
}
\maketitle

\begin{abstract}

Long Short-Term Memory (LSTM) is widely used in speech recognition. In order to achieve higher prediction accuracy, machine learning scientists have built increasingly larger models. Such large models are both computation and memory intensive. Deploying such bulky models results in high power consumption and leads to a high total cost of ownership (TCO) for a data center. 

To speedup prediction and make it energy efficient, we first propose a \emph{load-balance-aware pruning} method that can compress the LSTM model size by 20$\times$ (10$\times$ from pruning and 2$\times$ from quantization) with negligible loss of prediction accuracy. Also we proposed load-balance-aware pruning to ensure high hardware utilization. Next, we propose a scheduler that encodes and partitions the compressed model to multiple PEs for parallelism and schedules the complicated LSTM data flow. Finally, we design a hardware architecture named ESE that works directly on the sparse LSTM model. 

Implemented on Xilinx XCKU060 FPGA running at 200MHz, ESE has a performance of 282 GOPS working directly on the sparse LSTM network, corresponding to 2.52 TOPS on the dense one, and processes a full LSTM for speech recognition with a power dissipation of 41 Watts. Evaluated on the LSTM for speech recognition benchmark, ESE is 43$\times$ and 3$\times$ faster than Core i7 5930k CPU and Pascal Titan X GPU implementations. It achieves 40$\times$ and 11.5$\times$ higher energy efficiency compared with the CPU and GPU respectively.

\end{abstract}


\section{Introduction}

Deep neural network is widely used for speech recognition\cite{deepspeech1,deepspeech2}. Long Short-Term Memory (LSTM) and Gated Recurrent Unit (GRU) are two popular types of recurrent neural networks (RNNs) used for speech recognition. In this work, we evaluated the most complex one: LSTM~\cite{lstm}. A similar methodology could be easily applied to other types of recurrent neural networks.

Despite its high prediction accuracy, LSTM is hard to deploy because of its high computation complexity and memory footprint, leading to high power consumption. Memory reference consumes more than two orders of magnitude more energy than ALU operations, thus we focus on optimizing the memory footprint.

\begin{figure}[t!]
\centering
\vspace{10pt}
\scalebox{1}[0.9]{\includegraphics[width=0.48\textwidth]{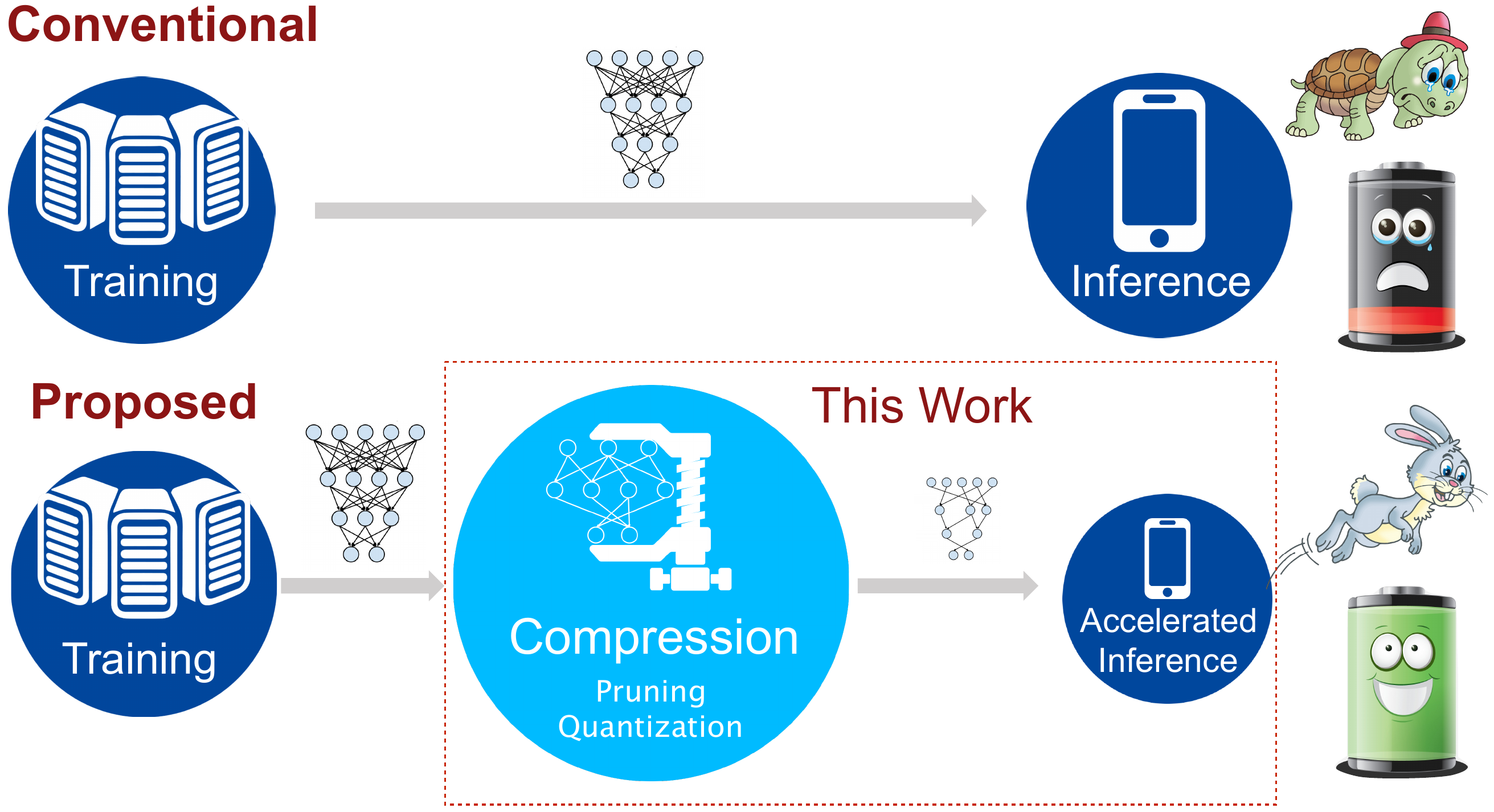}}
\vspace{-10pt}
\caption{Proposed efficient DNN deployment flow: model compression+accelerated inference.}
\label{fig:flow}
\vspace{-5pt}
\end{figure}

\begin{figure}[t!]
\centering
% \vspace{-10pt}
\scalebox{1}[0.9]{\includegraphics[width=0.48\textwidth]{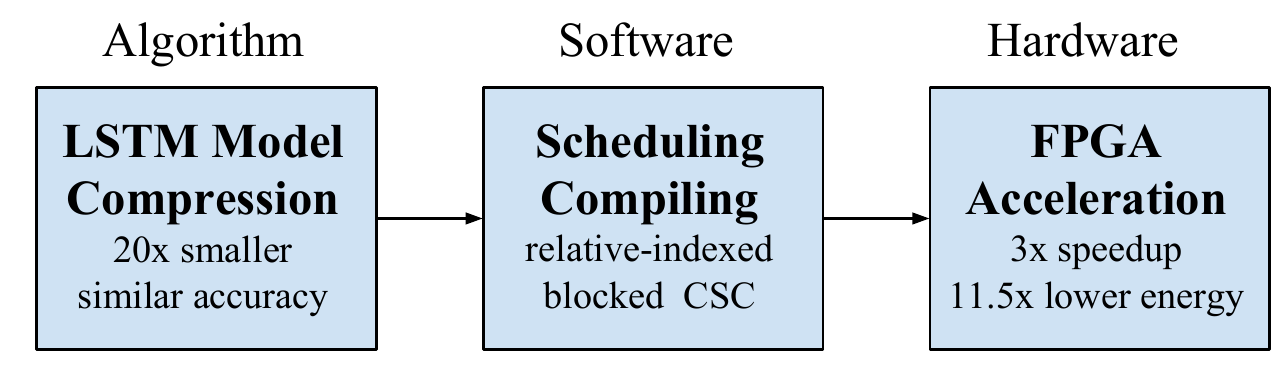}}
\vspace{-10pt}
\caption{ESE optimizes LSTM computation across algorithm, software and hardware stack.}
\label{fig:overview}
\vspace{-5pt}
\end{figure}

To reduce the memory footprint, we design a novel method to optimize across the algorithm, software and hardware stack: we first optimize the algorithm by compressing the LSTM model to 5\% of it's original size (10\% density and 2$\times$ narrower weights) while retaining similar accuracy; then we develop a software mapping strategy to represent the compressed model in a hardware-friendly way; finally we design specialized hardware to work directly on the compressed LSTM model. 

The proposed flow for efficient deep learning inference is illustrated in Fig.~\ref{fig:flow}. It shows a new paradigm for efficient deep learning inference, from \emph{Training=>Inference}, to \emph{Training=>Compression=>Accelerated Inference}, which has advantage of faster inference speed and energy efficiency compared with the conventional method. Using LSTM as a case study for the proposed paradigm, the design flow is illustrated in Fig.~\ref{fig:overview}.

The main contributions of this work are
\begin{enumerate}
\item We present an effective model compression algorithm for LSTM, which is composed of pruning and quantization.  We highlight our load-balance-aware pruning and automatic flow for dynamic-precision data quantization.

\item The recurrent nature of RNN and LSTM produces complicated data dependency, which is more challenging than feedforward neural nets. We design a scheduler that can efficiently schedule the complex LSTM operations with memory reference overlapped with computation. 

\item The irregular computation pattern after compression posed a challenge to hardware. We design a hardware architecture that can work directly on the sparse model. ESE achieves high efficiency by load balancing and partitioning both the computation and storage. ESE also supports processing multiple user's speech data concurrently. 

\item We present an in-depth study of the LSTM and speech recognition system and  optimize across the algorithm, software, hardware boundary. We jointly analyze the trade-off between prediction accuracy and prediction latency.

\end{enumerate}

\section{Background}

\begin{figure}[t]
\centering
\vspace{-5pt}
\scalebox{1}[0.95]{\includegraphics[width=0.45\textwidth]{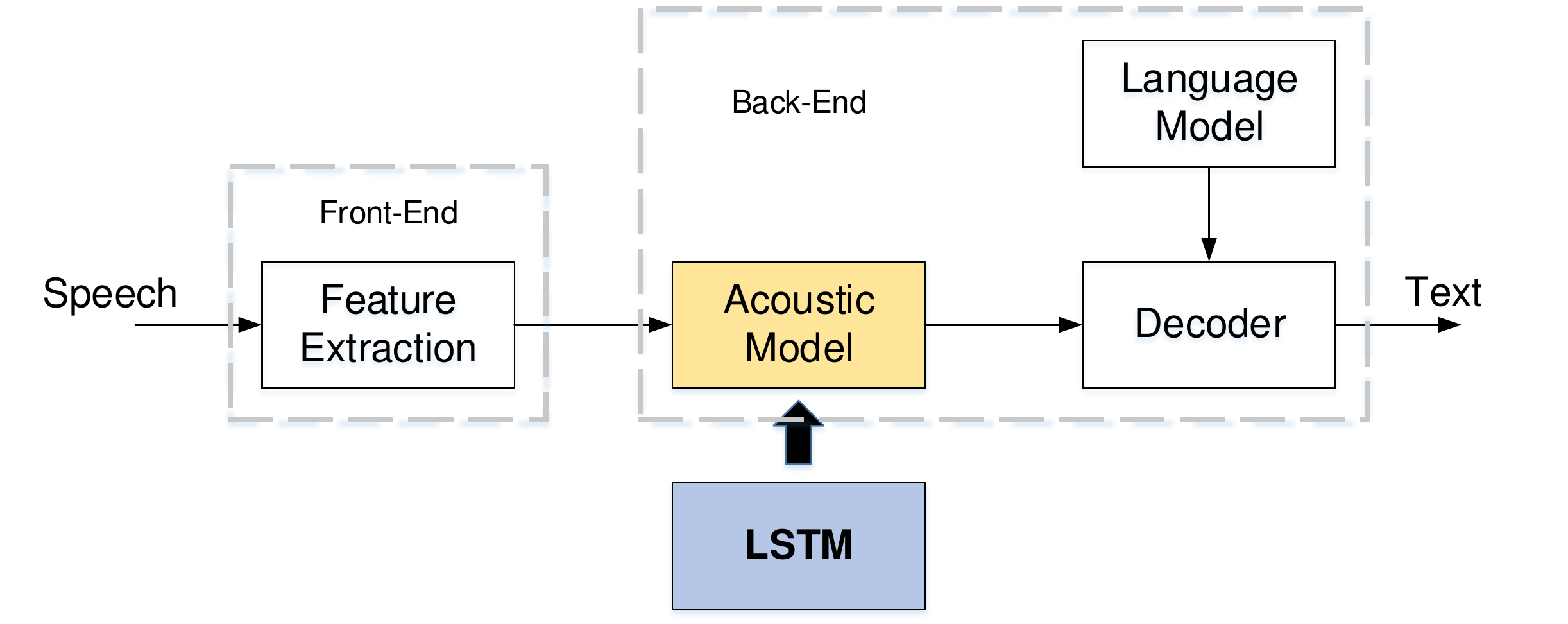}}
\vspace{-5pt}
\caption{The speech recognition pipeline. LSTM takes more than 90\% of the total execution time in the whole computation pipeline.}
\label{fig:asr}
\vspace{-5pt}
\end{figure}

Speech recognition is the process of converting speech signals to a sequence of words. As shown in Fig.~\ref{fig:asr}, the speech recognition system contains the front-end and back-end units, where the front-end unit is used for extracting features from speech signals, and the back-end unit processes the features and converts speech to text. The back-end includes an acoustic model (AM), language model (LM), and decoder. Here, the Long Short-Term Memory (LSTM) recurrent neural network is used in the acoustic model.

The feature vectors extracted from the front-end unit are processed by the acoustic model; then the decoder uses both acoustic and language models to generate the sequence of words by maximum a posteriori probability (MAP) estimation, which can be described as
\begin{equation*}
\hat{\mathbf{W}} = \underset{\mathrm{W}}{\mathrm{arg\,max}}\,P(\mathbf{W}|\mathbf{X}) = \underset{\mathrm{W}}{\mathrm{arg\,max}}\,\frac{P(\mathbf{X}|\mathbf{W})P(\mathbf{W})}{P(\mathbf{X})}
\end{equation*}
where for the given feature vector $\mathbf{X} = X_1 X_2 \ldots X_n$, the goal of speech recognition is to find the word sequence $\hat{\mathbf{W}} = W_1 W_2 \ldots W_m$ with maximum posterior probability $P(\mathbf{W}|\mathbf{X})$. Because $\mathbf{X}$ is fixed, the above equation can be rewritten as
\begin{equation*}
\hat{\mathbf{W}} = \underset{\mathrm{W}}{\mathrm{arg\,max}}\,P(\mathbf{X}|\mathbf{W})P(\mathbf{W})
\end{equation*}
where $P(\mathbf{X}|\mathbf{W})$ and $P(\mathbf{W})$ are the probabilities computed by acoustic and language models shown respectively in Fig.~\ref{fig:asr}\cite{sr-overview-msra}.

In modern speech recognition system, LSTM architecture is often used in large-scale acoustic modeling and for computing acoustic output probabilities. LSTM is the most computation and memory intensive part of the speech recognition pipeline. Thus we focus on accelerating the LSTM.

\begin{figure}[t!]
% \vspace{-10pt}
\centering
\scalebox{1}[0.85]{\includegraphics[width=0.45\textwidth]{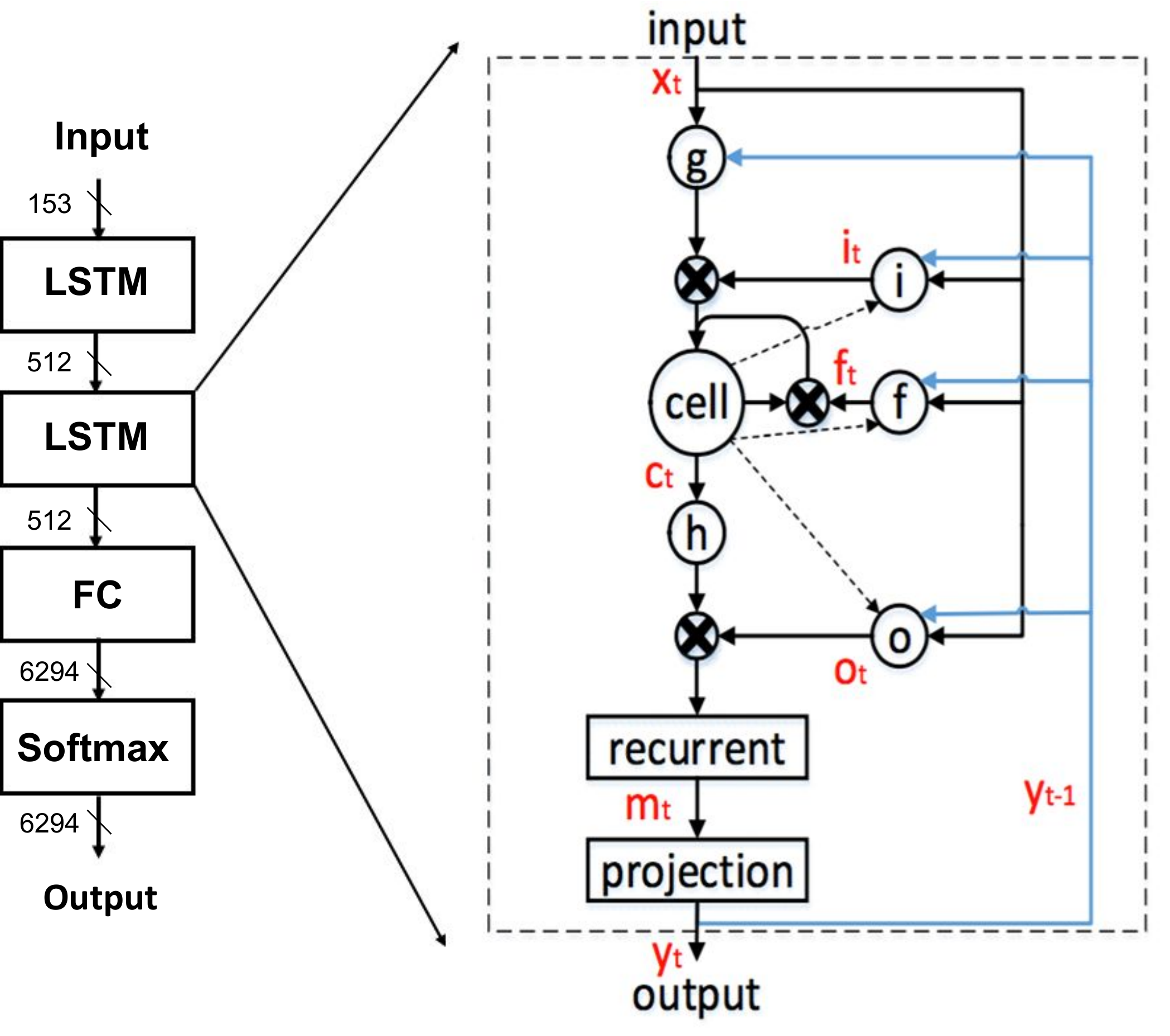}}
\vspace{-10pt}
\caption{Data flow of the LSTM model.}
\label{fig:lstm_arch}
\vspace{-15pt}
\end{figure}

The LSTM architecture is shown in Fig.~\ref{fig:lstm_arch}, which is the same as the standard LSTM implementation~\cite{sak2014long}. LSTM is one type of RNN, where the input at time $T$ depends on the output at $T-1$. Compared to the traditional RNN, LSTM contains special memory blocks in the recurrent hidden layer. The memory cells with self-connections in memory blocks can store the temporal state of the network. The memory blocks also contain special multiplicative units called gates: input gate, output gate and forget gate. As in Fig.~\ref{fig:lstm_arch}, the input gate $i$ controls the flow of input activations into the memory cell. The output gate $o$ controls the output flow into the rest of the network. The forget gate $f$ scales the internal state of the cell before adding it as input to the cell, which can adaptively forget the cell's memory.

An LSTM network accepts an input sequence $x = (x_1; \ldots; x_T )$, and computes an output sequence $y = (y_1; \ldots; y_T)$ by using the following equations iteratively from $t = 1 \text{ to } T$:
\begin{align}
i_t &=  \sigma(W_{ix}x_t + W_{ir}y_{t-1} + W_{ic}c_{t-1} + b_i) \\
f_t &=  \sigma(W_{fx}x_t + W_{fr}y_{t-1} + W_{fc}c_{t-1} + b_f) \\
g_t &=  \sigma(W_{cx}x_t + W_{cr}y_{t-1} + b_c) \\
c_t &= f_t \bigodot c_{t-1} + g_t \bigodot i_t \\
o_t &= \sigma(W_{ox}x_t + W_{or}y_{t-1} + W_{oc}c_t + b_o) \\
m_t &= o_t \bigodot h(c_t) \\
y_t &= W_{ym}m_t
\end{align}
Here the big O dot operator means element-wise multiplication, the $W$ terms denote weight matrices (e.g. $W_{ix}$ is the matrix of weights from the input to the input gate), and $W_{ic}$, $W_{fc}$, $W_{oc}$ are diagonal weight matrices for peephole connections. The $b$ terms denote bias vectors, while $\sigma$ is the logistic sigmoid function. The symbols $i$, $f$, $o$, $c$ and $m$ are respectively the input gate, forget gate, output gate, cell activation vectors and cell output activation vectors, and all of which are the same size. The symbols $g$ and $h$ are the cell input and cell output activation functions.

\section{Model Compression}
It has been widely observed that deep neural networks usually have a lot of redundancy~\cite{song_pruning,han2015deep}. Getting rid of the redundancy doesn't hurt prediction accuracy. From the hardware perspective, model compression is critical for saving the computation as well as the memory footprint, which means lower latency and better energy efficiency. We'll discuss two steps of model compression that consist of pruning and quantization in the next three subsections.

\subsection{Pruning}

In the pruning phase we first train the model to learn which weights are necessary, then prune away weights that are not contributing to the prediction accuracy; finally, we retrain the model given the sparsity constraint. The process is the same as \cite{song_pruning}. In step two, the saliency of the weight is determined by the weight's absolute value: if the weight's absolute value is smaller than a threshold, then we prune it away. The pruning threshold is empirical: pruning too much will hurt the accuracy while pruning at the right level won't. 

Our pruning experiments are performed on the Kaldi speech recognition toolkit\cite{povey2011kaldi}.  The trade-off curve of the percentage of parameters pruned away and phone error rate (PER)  is shown in Fig.\ref{fig:accuracy}. The LSTM is evaluated on the TIMIT dataset~\cite{timit}. Not until we prune away more than 93\% of parameters did the PER begin to increase dramatically. We further experimented on a proprietary dataset that is much larger: it has 1000 hours of training speech data, 100 hours of validation speech data, and 10 hours of test speech data. We find that we can prune away 90\% of the parameters without hurting word error rate (WER), which aligns with our result on the TIMIT dataset. In our later discussions, we use 10\% density (90\% sparsity).

\subsection{Load Balance-Aware Pruning}
On top of the basic deep compression method, we highlight our practical design considerations for hardware efficiency. To execute sparse matrix multiplication in parallel, we propose the load-balance-aware pruning method, which is very critical for better load balancing and higher utilization on the hardware.

Pruning could lead to a potential problem of unbalanced non-zero weights distribution. The workload imbalance over PEs may cause a gap between the real performance and peak performance. This problem is further addressed in Section~\ref{encode}.

Load-balance-aware pruning is designed to solve this problem and obtain hardware-friendly sparse network, which produces the same sparsity ratio among all the submatrices. During pruning, we make efforts to avoid the scenario when the density of one submatrix is 5\%  while the other is 15\%. Although the overall density is about 10\%, the submatrix with a density of 5\% has to wait for the other one with more computation, which leads to idle cycles. Load-balance-aware pruning assigns the same sparsity quota to submatrices, thus ensures an even distribution of non-zero weights.

\begin{figure}[t]
\centering
\hspace{-10pt}
\scalebox{1}[1]{\includegraphics[width=0.49\textwidth]{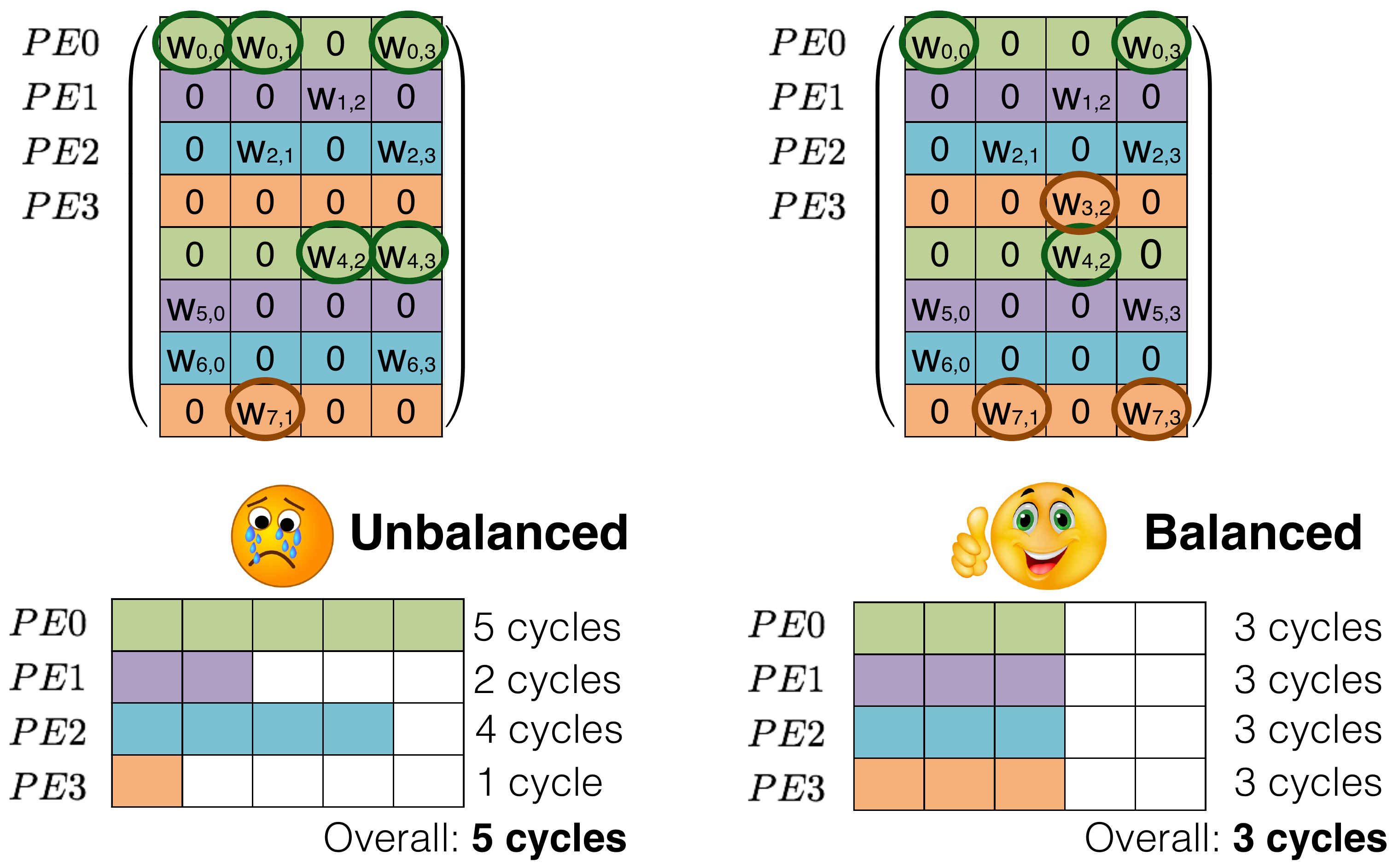}}
%\vspace{-5pt}
\caption{Load Balance Aware Pruning and its Benefit for Parallel Processing}
\label{fig:loadbalance}
%\vspace{-15pt}
\end{figure}

\begin{figure}[t]
\centering
\vspace{-5pt}
\scalebox{1}[1]{\includegraphics[width=0.48\textwidth]{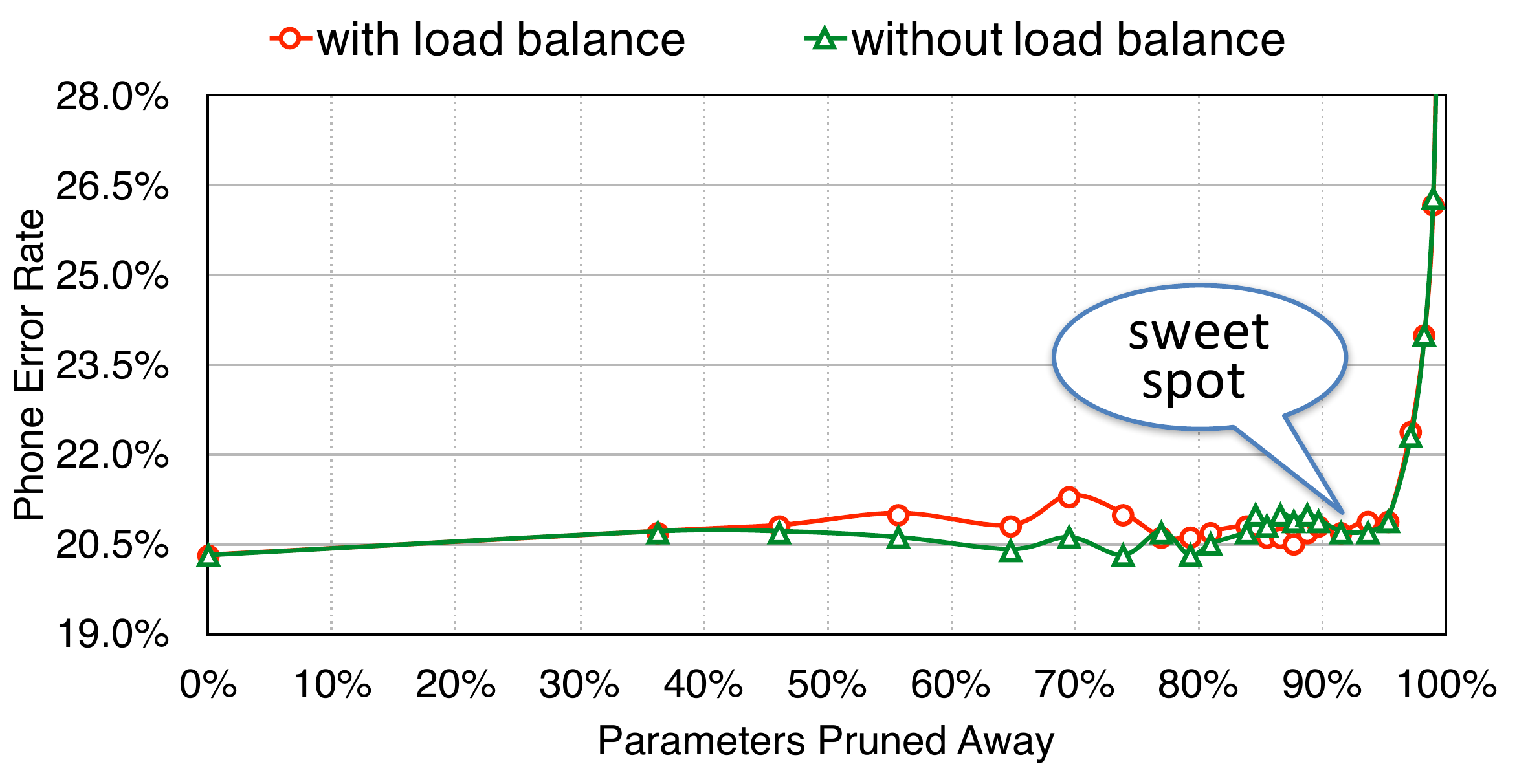}}
\vspace{-10pt}
\caption{Accuracy curve of load-balance-aware pruning and original pruning.}
\label{fig:accuracy}
\vspace{-5pt}
\end{figure}

In Fig.~\ref{fig:loadbalance}, the matrix is divided into four colors, and each color belongs to a PE for parallel processing. With conventional pruning, $PE_0$ might have five non-zero weights while $PE_3$ may have only one. The total processing time is restricted to the longest one, which is five cycles. With load-balance-aware pruning, all PEs have three non-zero weights; thus only three cycles are necessary to carry out the operation. Both cases have the same non-zero weights in total, but load-balance-aware pruning needs fewer cycles. The difference of prediction accuracy with/without load-balance-aware pruning is very small, as shown in Fig.~\ref{fig:accuracy}. There is some noise around 70\% sparsity, so we focused our experiments around 90\% sparsity, which is the sweet spot. We find the performance is very similar. 

To show that load-balance-aware pruning still obtains comparable prediction accuracy, we compare it with original pruning on the TIMIT dataset. As demonstrated in Fig.\ref{fig:accuracy}, the accuracy margin between two methods is within the variance of pruning process itself.

\subsection{Weight and Activation Quantization}
We further compressed the model by quantizing 32bit floating point weights into 12bit integer. We used linear quantization strategy on both the weights and activations.

In the weight quantization phase, the dynamic ranges of weights for all matrices in each LSTM layer are analyzed first, then the length of the fractional part is initialized to avoid data overflow. 

The activation quantization phase aims to figure out the optimal solution to the activation functions and the intermediate results. We built lookup tables and use linear interpolation for the activation functions, such as sigmoid and tanh, and analyze the dynamic range of their inputs to decide the sampling strategies. We also investigated the minimum amount of bits to maintain the accuracy. 

We explored different data quantization strategies with LSTM trained under TIMIT corpus. Performing the weight and activation quantization, we can achieve 12bit quantization without any accuracy loss. The data quantization strategies are shown in Table .\ref{tab:weight}, \ref{tab:function}, \ref{tab:activation}. For the lookup tables of activation functions sigmoid and tanh, the sampling ranges are [-64, 64] and [-128, 128] respectively. The sampling points are both 2048, and the outputs are 16bit with 15bit decimals. All the results are obtained using the Kaldi framework.
 
\begin{table}[t]
\vspace{-10pt}
\caption{Weight Quantization under different Bits.}
\centering
\scriptsize
\setlength{\tabcolsep}{3pt}
\begin{threeparttable}
\begin{tabular}{|c|c|r|r|c|c|c|c|}
\hline
%\multicolumn{2}{|c|}{\multirow{2}{*}{Weight Matrices}\tnote{1}} &\multirow{2}{*}{Min} &\multirow{2}{*}{Max} &\multirow{2}{*}{Integer} &\multicolumn{3}{c|}{Decimals}\\
\multicolumn{2}{|c|}{\multirow{2}{*}{Weight Matrices\tnote{1}}} & \multicolumn{1}{c|}{\multirow{2}{*}{Min}} & \multicolumn{1}{c|}{\multirow{2}{*}{Max}} &\multirow{2}{*}{Integer} &\multicolumn{3}{c|}{Decimals}\\
\cline{6-8}
\multicolumn{2}{|c|}{} & & & & 16bit &12bit &8bit \\
\hline
\multirow{7}{*}{LSTM1} &W\_gifo\_x\tnote{2} &-4.9285 &5.7196 &4 &8 &4 &0\\
\cline{2-8}
&W\_gifo\_r\tnote{2}    &-0.6909    &0.7140        &1    &11    &7    &3\\
\cline{2-8}
&bias        &-3.0143    &2.1120        &3    &13    &9    &5\\
\cline{2-8}
&W\_ic        &-0.6884    &0.9584        &1    &15    &11    &7\\
\cline{2-8}
&W\_fc        &-0.6597    &0.7204        &1    &15    &11    &7\\
\cline{2-8}
&W\_oc        &-1.5550    &1.3325        &2    &14    &10    &6\\
\cline{2-8}
&W\_ym        &-0.9373    &0.8676        &1    &11    &7    &3\\
\hline
\multirow{7}{*}{LSTM2} &W\_gifo\_x &-1.0541 &1.0413 &2 &10 &6 &2\\
\cline{2-8}
&W\_gifo\_r    &-0.6313    &0.6400        &1    &11    &7    &3\\
\cline{2-8}
&bias        &-1.5833    &1.8009        &2    &14    &10    &6\\
\cline{2-8}
&W\_ic        &-0.9428    &0.5158        &1    &15    &11    &7\\
\cline{2-8}
&W\_fc        &-0.5762    &0.6202        &1    &15    &11    &7\\
\cline{2-8}
&W\_oc        &-1.0619    &1.4650        &2    &14    &10    &6\\
\cline{2-8}
&W\_ym        &-1.0947    &1.0170        &2    &10    &6    &2\\
\hline
\end{tabular}
\begin{tablenotes}
    \item[1] Only weights in LSTM layers are qunantized.
    \item[2] In Kaldi, Wcx, Wix, Wfx, Wox are saved together as W\_gifo\_x, and so does W\_gifo\_r mean.
\end{tablenotes}
\label{tab:weight}
\end{threeparttable}
\vspace{-15pt}
\end{table}

\begin{table}[t]
\caption{Activation Function Lookup Table.}
\centering
\scriptsize
\setlength{\tabcolsep}{3pt}
\begin{tabular}{|c|c|c|c|c|}
\hline
Activation        &Min        &Max        &sampling range    &sampling points\\
\hline
Sigmoid Input    &-51.32        &59.16        &-64-64            &2048\\
\hline
Tanh Input        &-104.7        &107.4        &-128-128        &2048\\
\hline
\end{tabular}
\label{tab:function}
\vspace{-10pt}
\end{table}

\begin{table}[t]
\vspace{-10pt}
\caption{Other Activation Quantization.}
\centering
\scriptsize
\begin{tabular}{|c|c|c|c|c|}
\hline
Activation                &Min        &Max        &Width    &Decimals\\
\hline
LSTM Input                &-7.611        &8.166        &16            &11\\
\hline
Intermediate Results    &-107.8        &109.4        &16            &8\\
\hline
\end{tabular}
\label{tab:activation}
\vspace{-10pt}
\end{table}

For TIMIT, as shown in Table .\ref{tab:WER}, the PER is 20.4\% for the original network and changes to 20.7\% after the pruning and fine-tune procedure when 32-bit floating-point numbers are used. The PER remains as 20.7\% without any accuracy loss under 16/12-bit quantization, and deteriorates to 84.5\% while 8-bit quantization is employed.

\begin{table}[t]
\caption{PER Before and After Compression.}
\vspace{5pt}
\centering
\begin{tabular}{|c|c|}
\hline
Quantization Scheme    &Phone Error Rate~\%\\
\hline
32bit floating original network    &20.4\%\\
\hline
32bit floating pruned network    &20.7\%\\
\hline
16bit fixed    pruned network        &20.7\%\\
\hline
\bf{12bit fixed pruned network}        &\bf{20.7\%}\\
\hline
8bit fixed    pruned network        &84.5\%\\
\hline
\end{tabular}
\label{tab:WER}
\vspace{-10pt}
\end{table}
% \vspace{0.3\textheight}
\section{Encoding and Compiling}
\label{encode}

\begin{figure}[b]
\centering
\vspace{-10pt}
\scalebox{1}[0.9]{\includegraphics[width=0.4\textwidth]{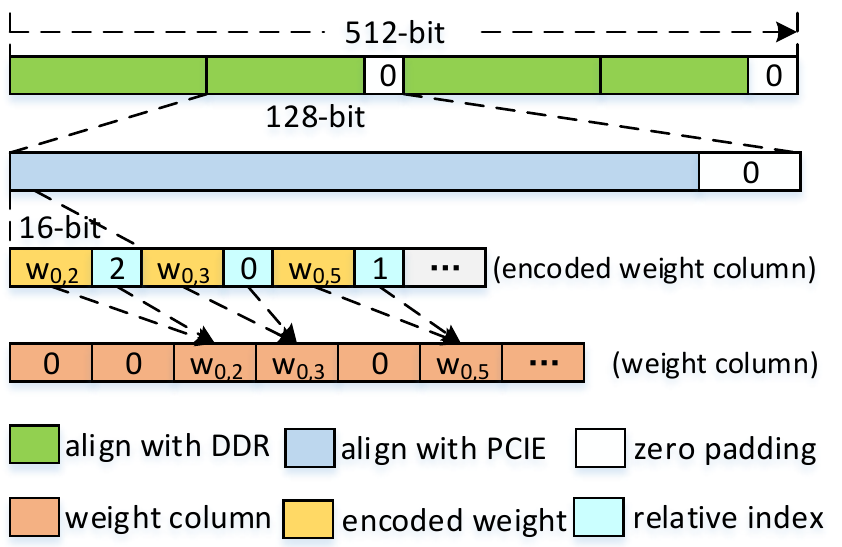}}
\vspace{-10pt}
\caption{Encoding in CSC format and data align using zero-padding. }
\label{fig:encoding}
\vspace{-15pt}
\end{figure}

\begin{figure}[b]
\centering
\vspace{-10pt}
\scalebox{1}[0.9]{\includegraphics[width=0.48\textwidth]{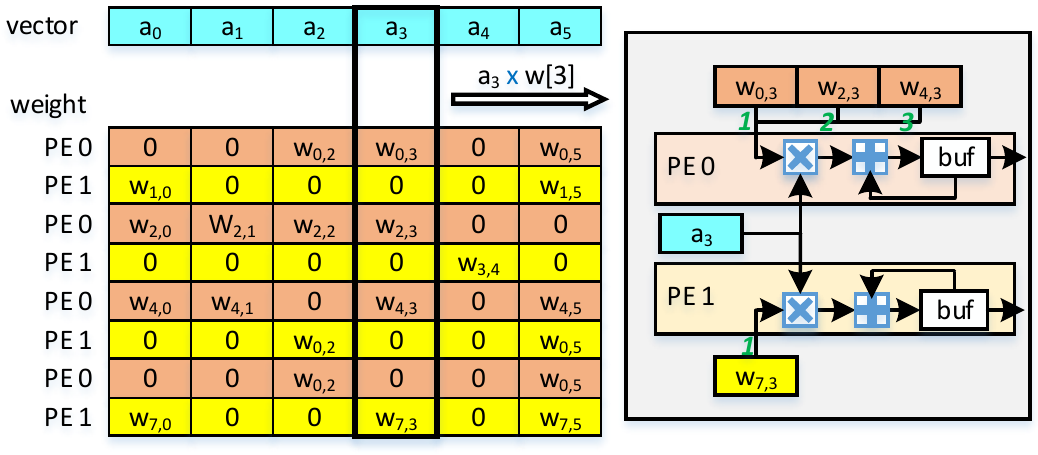}}
\vspace{-10pt}
\caption{The computation pattern: non-zero weights in a column are assigned to 2 PEs, and every PE multiply-add their weights with the same element from the shared vector.}
\label{fig:ComputationPattern}
\vspace{-15pt}
\end{figure}

The LSTM computation includes sparse matrices multiplication, element-wise multiplication, and memory reference. We designed a data flow scheduler to make full use of the hardware accelerator.

Data is divided into $n$ blocks by row where $n$ is the number of PEs in one channels of our hardware accelerator. The first $n$ rows are put in $n$ different PEs. The $n+1$ row is put in the first PE again. This ensures that the first part of the matrix will be read in the first reading cycle and can be used in the next step computation immediately.

Because of the sparsity of pruned matrices. We only store the nonzero number in weight matrices to save redundant memory. We use relative row index and column pointer to help store the sparse matrix. The relative row index for each weight shows the relative position from the last nonzero weight. The column pointer indicates where the new column begins in the matrix. The accelerator will read the weight according to the column pointer.

Considering the byte-aligned bit width limitation of DDR, we use 16bit data to store the weight. The quantized weight and relative row index are put together (i.e. 12bit for quantized weight and 4bit for relative row index).

Fig.\ref{fig:encoding} shows an example for the compressed sparse column (CSC) storage format and zero-padding method. We locate one column in the weight matrix through a pointer and calculate the absolute address of weights by accumulating relative indexes.  In Fig. \ref{fig:ComputationPattern}, we demonstrate the computation pattern using a simple example where the input vector has 6 elements \{$a_{0}$,$a_{1}$,$a_{2}$,$a_{3}$,$a_{4}$,$a_{5}$\}, and the weight matrix contains 8$\times$6 elements. There are 2 PEs calculating $a_{3}$$\times$$w[3]$, where $a_{3}$ is the fourth element in the input vector and $w[3]$ represents the fourth column in the weight matrix.

% \vspace{0.3\textheight}
\section{Hardware Implementation}

In this section, we first present challenges in hardware design and then propose the Efficient Speech Recognition Engine (ESE) accelerator system and detail how ESE accelerates the sparse LSTM.

% \newpage
\subsection{Motivation}

Although pruning and quantization can reduce the memory footprint, three new challenges are introduced. General purpose processors cannot implement these challenges efficiently.

First, irregular computation is introduced by compression. After pruning, dense computation becomes sparse computation; After quantization, the weight and index are not byte-aligned and must be grouped. We group the 4-bit pointer, and 12-bit weight into 2 bytes.

\begin{figure}[t]
\centering
\vspace{-15pt}
\scalebox{1}[0.9]{\includegraphics[width=0.4\textwidth]{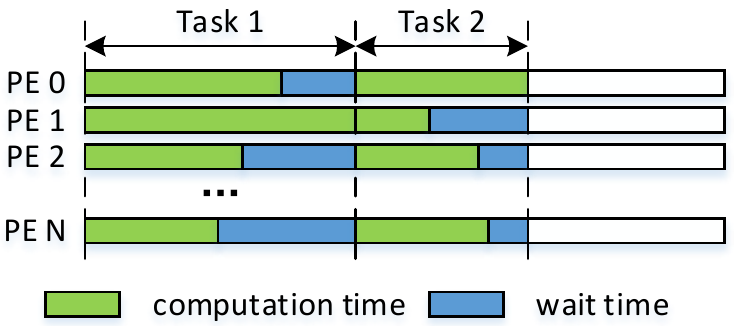}}
\vspace{-10pt}
\caption{Imbalanced workload results in more waiting time.}
\label{fig:unbalance}
\vspace{-15pt}
\end{figure}

Second, load imbalance introduced by sparsity will reduce the hardware efficiency. In the sparse LSTM, a single element in the voice vector will be consumed by multiple PEs. As a result, operations of all PEs have to be synchronized. It will create a long waiting period if some PEs have fewer non-zero weights, as shown in Fig.\ref{fig:unbalance}. 

Moreover, general-purpose processors cannot fully exploit the parallelism in the compressed LSTM network. In the custom design, however, we have the freedom to take advantage of the parallelism of both the inter sparse SpMV operation and the intra SpMV operation.

Many challenges exist in the specialized hardware accelerator design on FPGA. 
First, customized decoding circuits are needed to recover the original weight matrix. The index is relative, so accumulation is needed to recover the absolute index. We use only 4-bits to represent relative offset. If a real offset is more than 16, the largest offset that 4 bits can represent, a padding zero is introduced. 

Second,  data representation should be carefully designed.
The data width of the PCIE interface, external DDR3 memory interface, and data itself are not aligned. 
Moreover, the dynamic-precision quantization makes hardware computation on different data more complex and irregular. Bit shifts are necessary for different layers. 

Third, a carefully designed scheduler/controller is needed. The LSTM network involves a complicated data flow and many different types of weights. Computations in the LSTM network have dependency on each other. Some computation can be executed concurrently, while other computation has to be executed sequentially.  Moreover, the hardware design should support input vector sharing in the multi-channel system, which aims to perform multiple LSTM networks with different voice vectors concurrently. Therefore, a carefully designed scheduler is necessary for a highly pipelined design, which can overlap the data communication and computation.

\subsection{System Overview}

\begin{figure*}[t!]
\centering
\vspace{-10pt}
\scalebox{1}[1]{\includegraphics[width=0.95\textwidth]{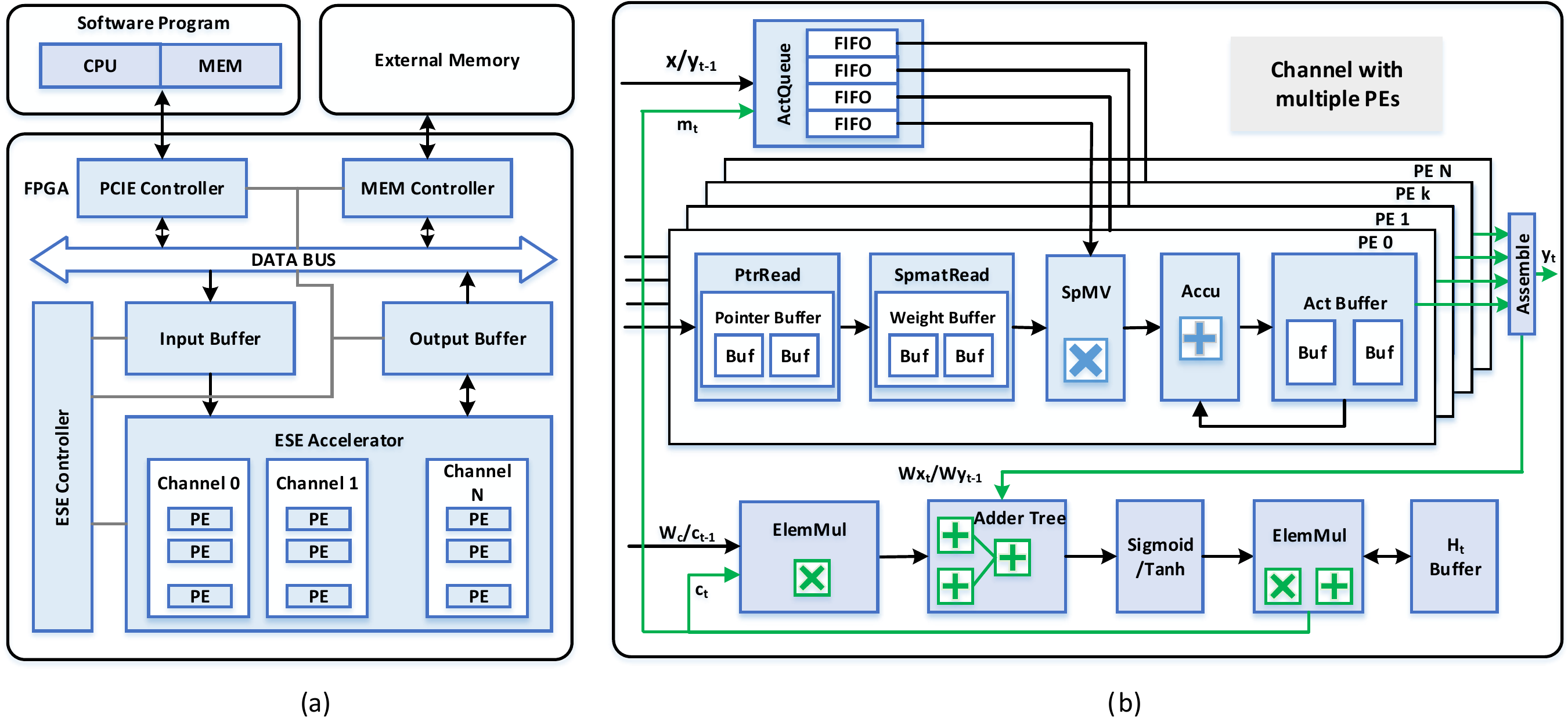}}
\vspace{-15pt}
\caption{The Efficient Speech Recognition Engine (ESE) accelerator system: (a) the overall ESE system architecture; (b) one channel with multiple processing elements (PEs).}
\label{fig:ESEOverview}
% \vspace{-5pt}
\end{figure*}

Fig.\ref{fig:ESEOverview} (a) shows the overview architecture of the ESE system. It is a CPU+FPGA heterogeneous architecture to accelerate LSTM. The whole system can be divided into three parts: the hardware accelerator on a FPGA chip, the software program on CPU, and the external memory on the FPGA board.

The software part consists of a CPU and host memory. It communicates with FPGA via the PCI-Express bus. In the initialization procedure, it sends parameters of the LSTM model to FPGA. It can transmit voice vectors and receive corresponding results if the hardware accelerator on FPGA is ready. 

The external memory together with the FPGA chip on one development board stores all the parameters and voice vectors. The on-chip BRAM is limited while the amount of data in the LSTM model is larger than it can hold. The accelerator accesses the DRAM through memory controller (MEM Controller), which is built using the memory interface generator (MIG) IP. 

On the FPGA chip, we put the ESE Accelerator, ESE Controller, PCIE Controller, MEM Controller, and On-chip Buffers. The ESE Accelerator consists of Processing Elements (PEs) which take charge of the majority of computation tasks in the LSTM model. PE is the basic computation unit for a slice of voice vectors with partial weight matrix. Each ESE channel implements the LSTM network for one voice vector sequence independently. 
On-chip buffers, including input buffer and output buffer,  prepare data to be consumed by PEs and store the generated results. The ESE Controller determines the behavior of other circuits on the FPGA chip. It schedules the PCIE/MEM Controller for data-fetch and the LSTM computation pipeline flow of the ESE Accelerator. The accelerator reads parameters and voice vectors from, and writes computation results to, the DRAM memory. When the MEM Controller is in the idle state, the accelerator can read results currently stored in the memory and feed them to the software part.

\begin{figure*}[h!]
\centering
% \vspace{10pt}
\scalebox{1}[1]{\includegraphics[width=0.95\textwidth]{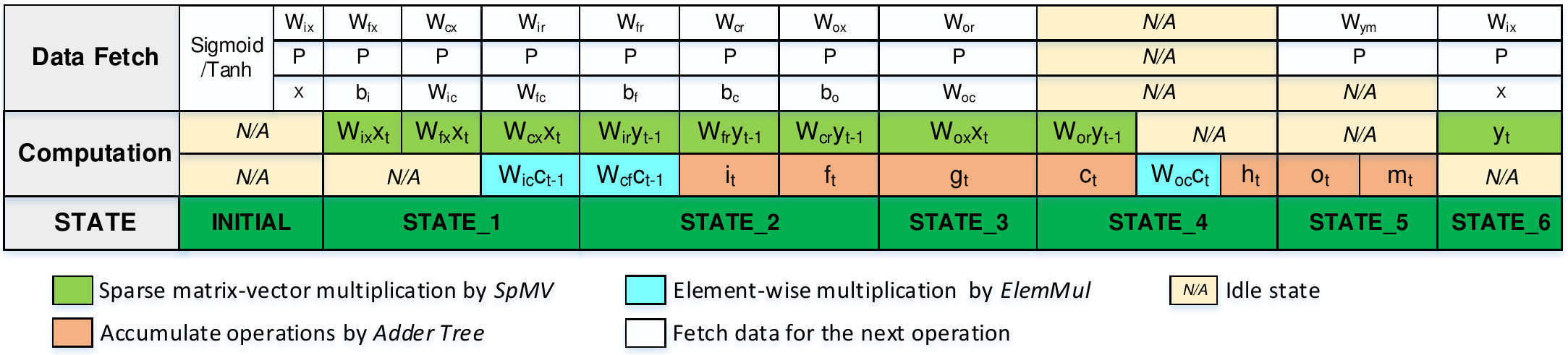}}
\vspace{-5pt}
\caption{The state flow of the ESE accelerator system: operations in the horizontal direction and vertical direction are executed sequentially and concurrently respectively.}
\label{fig:fsm}
\vspace{-5pt}
\end{figure*}

\subsection{ESE Controller (Scheduler)}

\begin{table}[t]
\vspace{-5pt}
\caption{Two types of LSTM operations: matrix-vector multiplication and element-wise multiplication.}
\centering
%\setlength{\tabcolsep}{3pt}
%\scriptsize
\begin{tabular}{|c|c|c|}
\hline
Target           &SpMV Group                            &ElemMul Group                  \\
\hline
$i_{t}$         &$W_{ix}$$x_{t}$, $W_{ir}$$y_{t-1}$    &$W_{ic}$$c_{t-1}$                   \\
\hline
$f_{t}$           &$W_{fx}$$x_{t}$, $W_{fr}$$y_{t-1}$    &$W_{fc}$$c_{t-1}$                   \\
\hline
$c_{t}$            &$W_{cx}$$x_{t}$, $W_{cr}$$y_{t-1}$    &$f_{t}$$c_{t-1}$, $i_{t}$$g_{t}$    \\
\hline
$o_{t}$            &$W_{ox}$$x_{t}$, $W_{or}$$y_{t-1}$    &$W_{oc}$$c_{t}$                     \\
\hline
$m_{t}$            &$N/A$                                  &$o_{t}$$h_{t}$                      \\
\hline
$y_{t}$            &$W_{ym}$$m_{t}$       &$N/A$                                    \\
\hline
\end{tabular}
\vspace{-5pt}
\label{tab:TaskPartition}
\end{table}

The most expensive operations are sparse matrix vector multiplication (SpMV) and element-wise multiplication (ElemMul). We partition the operations involved in the LSTM network described by equations (1) to (6), into the such two operations, as shown in Table~\ref{tab:TaskPartition}.

LSTM is a complicated dataflow. We want to meet the data dependency and ensure more parallelism at the same time. Fig.\ref{fig:fsm} shows the state machine in the ESE scheduler.  It overlaps computation and memory reference. From state INITIAL to STATE\_6, the ESE accelerator completes the computation of a LSTM. The first three lines operations are fetching weights, pointers, and vectors/diagonal matrix/bias respectively to prepare for the next computation. Operations in the fourth line are matrix-vector multiplications, and in the fifth line are element-wise multiplications (indigo blocks) or accumulations (orange blocks). Operations in the horizontal direction have to be executed sequentially, while those in the vertical direction can be executed concurrently.  For example, we can calculate $W_{fr}$$y_{t-1}$ and ${i_{t}}$ concurrently, because the two operations are not dependent on each other in the LSTM network, and they can be executed by two independent computation units. $W_{ir}$$y_{t-1}$/$W_{ic}$$c_{t-1}$ and ${i_{t}}$ have to be executed sequentially, because ${i_{t}}$ is dependent on the former operations in LSTM network. 

$W_{ix}$$x_{t}$ and $W_{fx}$$x_{t}$ are not dependent on each other in the LSTM network, but they cannot be calculated concurrently because they have resource conflict.  Weights are stored in one piece of DDR3 memory because even after compression the real world network cannot fit in the limited block RAM (4.25MB). Other parameters and input vector are stored in the other piece of DDR3 memory. Pointers are required for the same computations as weights, because we use pointers to look up weights in the compressed LSTM network. But the memory overhead necessary to store the pointers is small. Note that $x$, bias $b$ and diagonal matrix $W_{c}$ are not accessed at the same time, and all these parameters have a relatively small quantity. Therefore, pointers, vectors, diagonal matrix and bias can be stored in the same external memory and prepared accordingly during weight fetching period. 

The latency of the element-wise operations and non-linear functions is not on the critical path. These operations are executed in parallel with the matrix-vector multiplication and weights-fetching.

\subsection{ESE Channel Architecture}

Fig.\ref{fig:ESEOverview} (b) shows the architecture of one ESE channel with multiple PEs. It is composed of Activation Queue (ActQueue), Sparse Matix-vector Multiplier (SpMV), Accumulator, Element-wise Multiplier (ElemMul), Adder Tree, Sigmoid/Tanh Units and local buffers.

\textbf{Activation Vector Queue (ActQueue).} ActQueue consists of several FIFOs. Each FIFO stores some elements of the input voice vector $a_{j}$ for each PE. ActQueue is shared by all the PEs in one channel, while each FIFO is owned by each PE independently. 

ActQueue is used for decoupling the imbalanced workload among different PEs. Load imbalance arises when the number of multiply accumulation operations performed by every PE is different, due to the imbalanced sparsity. Those PEs with fewer computation tasks have to wait until the PE with the most computation tasks finishes. Thus if we have a FIFO, the fast PE can fetch a new element from the FIFO and won't need to be blocked by slow PEs. The data width of FIFO is 16-bit, the depth is adjusted from 1 to 16 to investigate its effects on the latency, and the results are discussed in the experiment section. These FIFOs are built on the distributed RAM on chip.

\textbf{Sparse Matrix Read (SpmatRead).} Pointer Read Unit (PtrRead) and Sparse Matrix Read (SpmatRead) manage the encoded weight matrix storage and output. The start and end pointers $p_{j}$ and $p_{j+1}$ for column j determine the start location and length of elements in one encoded weight column that should be fetched for each element of a voice vector. SpmatRead uses pointers $p_{j}$ and $p_{j+1}$ to look up the non-zero elements in weight column $j$. Both PtrRead and SpmatRead consist of ping-pong buffers. Each buffer can store 512 16-bit values and is implemented with block rams. Each 16-bit data in SpmatRead buffers consists of a 4-bit index and a 12-bit weight. Here are the four basic components.

\textbf{Sparse Matrix-vector Multiplication (SpMV).}  Each element in the voice vector is multiplied by its corresponding weight column. Multiplication results in the same row of all new vectors are summed to generate an element in the result vector, which is a local reduction.
In ESE, $SpMV$ multiplies an element from the input activation by a column of weight, and the current partial result is written into the partial result buffer $ActBuffer$. Accumulator $Accu$ sums the new output of $SpMV$ and previous data stored in Act Buffer. The multiplier instantiated in the design can perform 16bit$x$12bit functions.

\textbf{Element-wise Multiplication (ElemMul).} $ElemMul$ in Fig.\ref{fig:ESEOverview} (b) generates one vector by consuming two vectors. Each element in the output vector is the element-wise multiplication of two input vectors. There are 16 multipliers instantiated for element-wise multiplications per channel. 

\textbf{Adder Tree.} $Adder Tree$ performs summation by consuming the intermediate data produced by other units or bias data from input buffer.

\textbf{Sigmoid/Tanh.} $Sigmoid and Tanh$ units are the non-linear modules applied as activation functions to some intermediate summation results.

Here we explain how ESE computes $i_{t}$. In the initial state, PE receives weight $W_{ix}$, pointers $P$ and voice vector $x$. Then SpMV calculates $W_{ix}$$X_{t}$ in the first phase of STATE\_1. $W_{ir}$$y_{t-1}$ and $W_{ic}$$c_{t-1}$ are generated by SpMV and ElemMul respectively in the first phase of STATE\_2. In the second phase of STATE\_2, Adder Tree accumulates these output and bias data from the input buffer and then the following non-linear activation function unit Sigmoid/Tanh produces intermediate data $i_{t}$. PE will fetch required parameters in the previous phase to overlap with the computation. The other LSTM network operations are similar. In Fig.\ref{fig:fsm}, either SpMV or ElemMul is in the idle state at some phases. This is because both matrix-vector multiplication and element-wise multiplication consume weight data, while PE cannot pre-fetch enough weight data for both computations in the period of one phase.

\subsection{Memory System}

\begin{figure}[tb]
\centering
\vspace{-15pt}
\scalebox{1}[1]{\includegraphics[width=0.45\textwidth]{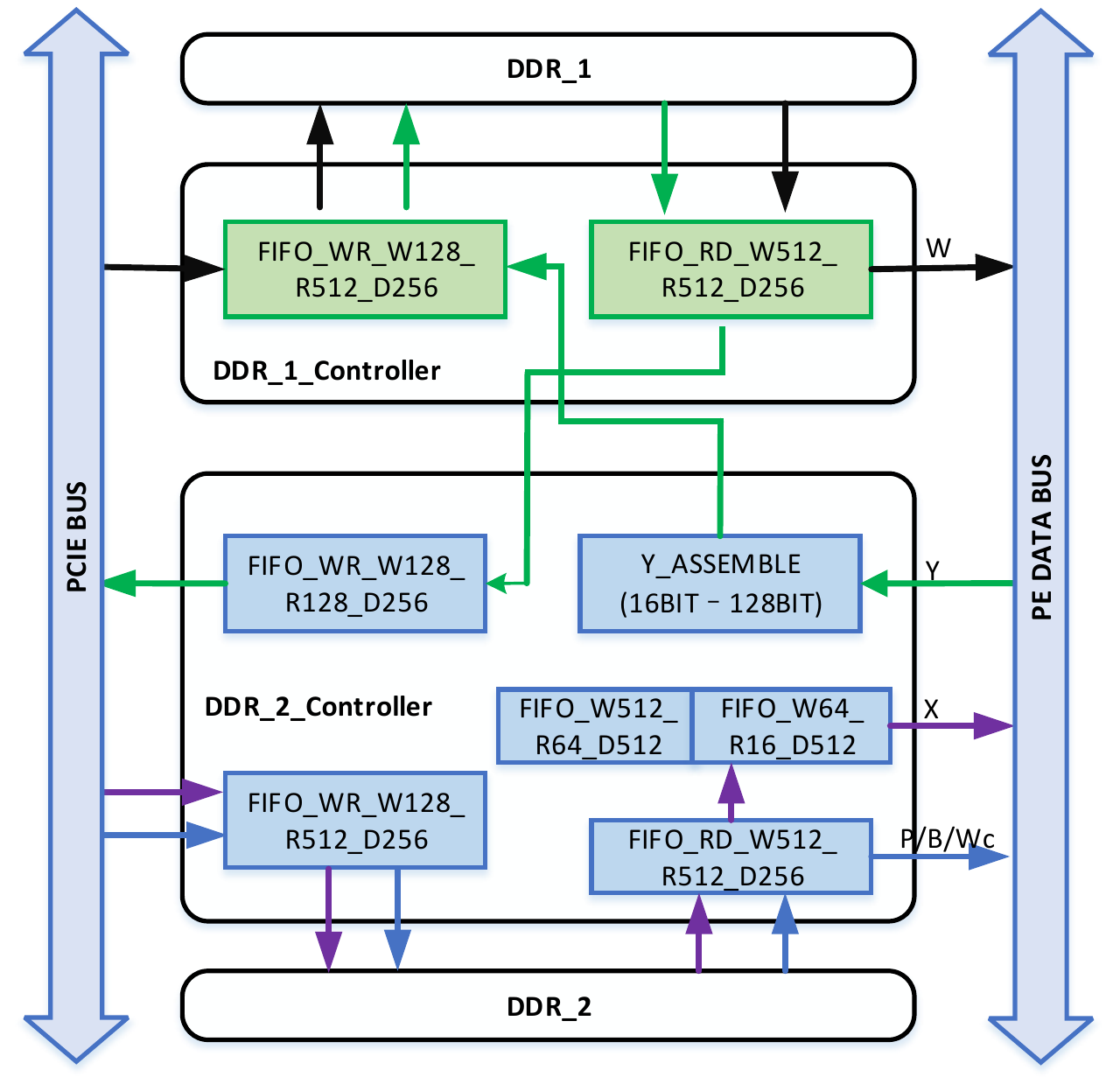}}
\vspace{-10pt}
\caption{Memory management unit.}
\label{fig:MEM}
\vspace{-15pt}
\end{figure}

In the hardware design, on-chip buffers are built upon a basic idea of double-buffering, in which double buffers are operated in a ping-pong manner to overlap data transfer with computation. We use two pieces of 4GB DDR3 DRAMs as the off-chip memory, named DDR\_1 and DDR\_2 in Fig.\ref{fig:MEM}, and design a memory controller (MEM Controller). Fig.\ref{fig:MEM} shows the MEM Controller architecture. On the one hand, it receives instructions from the ESE Controller and schedules the data flow among the ESE accelerator, PCIE interface, and DDR3 interface. On the other hand, it rearranges received data into structures required by the destination interface. We take the data flow of result $y$ as an example. Data $y$ at the output port of PE is 16-bit wide, while the PCIE interface is 128-bit wide. In order to increase the data transmission speed, we assemble eight 16-bit data into one 128-bit value by Y\_ASSEMBLE unit. Then the value will be stored in DDR\_1 temporarily and fed back to the software via PCIE interface when both PCIE and DDR\_1 are in idle state. The behavior described above is shown as the green arrow line in Fig.\ref{fig:MEM}. Similarly, vector $x$ is split into 32 16-bit values from a 512-bit value through asynchronous FIFOs. Moreover, asynchronous FIFOs, FIFO\_WR\_XX and FIFO\_RD\_XX also play an important role of asynchronous clock domains isolation.

\section{Experimental Results}

In this section, the performance of the hardware system is evaluated. First, we introduce the environment setup of our experiments. Then, hardware resource utilization and comprehensive experimental results are provided.

\subsection{Experimental Setup}

The proposed ESE hardware system is built on XCKU060 FPGA running at 200 MHz. Two external 4GB DDR3 DRAMs are used.
Our host program is responsible for sending parameters and vectors into the programmable logic part, and collecting corresponding results.

We use the TIMIT dataset to evaluate the performance of model compression. TIMIT is an acoustic-phonetic continuous speech corpus. It contains broadband recordings of 630 speakers of eight major dialects of American English, each reading ten phonetically rich sentences. We also use a proprietary, much larger speech recognition dataset that contains 1000 hours of training data, 100 hours of validation data and 10 hours of test data. 

Our baseline software program runs on i7-5930k CPU and Pascal Titan X GPU. We use MKL BLAS / cuBLAS on CPU / GPU for dense matrix operation implementations, and MKL SPARSE / cuSPARSE on CPU / GPU for sparse matrix implementations.

\subsection{Resource Utilization}
Table \ref{tab:ResourceUtilization} shows the resource utilization for our ESE design configured with 32 channels, and each channel has 32 PEs on XCKU060 FPGA. The ESE accelerator design almost fully utilizes the FPGA's hardware resource.

We configured each channel with 32 PEs, which is determined by balancing computation and data transfer. It is required that the speed of data transfer is no less than that of computation in order not to starve the DSP. As a result, we get Equation \ref{eq:PE_NUM_CAL}. The expression to the left of the equal sign means that the amount of computations is divided by the computation speed. Multiplied by 2 in the numerator part means each piece of data needs multiplication and accumulation operations, and that in the denominator part indicates twice multiply-accumulate operations for 2 bytes (16-bit). ESE implements the multiply-accumulate operation in a pipeline manner. The expression to the right represents the cycles that ESE fetch the required amount of data from external memory. In our hardware implementation, both the frequencies of PE and memory interface controller are 200MHz. The width of external DRAM is 512-bit. Therefore, the proper number of PEs per channel is 32.

\begin{equation}
\begin{aligned}
&\frac{data\_size\times compress\_ratio\times 2}{PE\_num\times 2\times freq\_PE} \\ &\geq \frac{data\_size\times compress\_ratio\times 16bit}{ddr\_width\times freq\_ddr}
\label{eq:PE_NUM_CAL}
\end{aligned}
\end{equation}

\begin{figure}[t]
\centering
% \vspace{-10pt}
\scalebox{1}[1]{\includegraphics[width=0.46\textwidth]{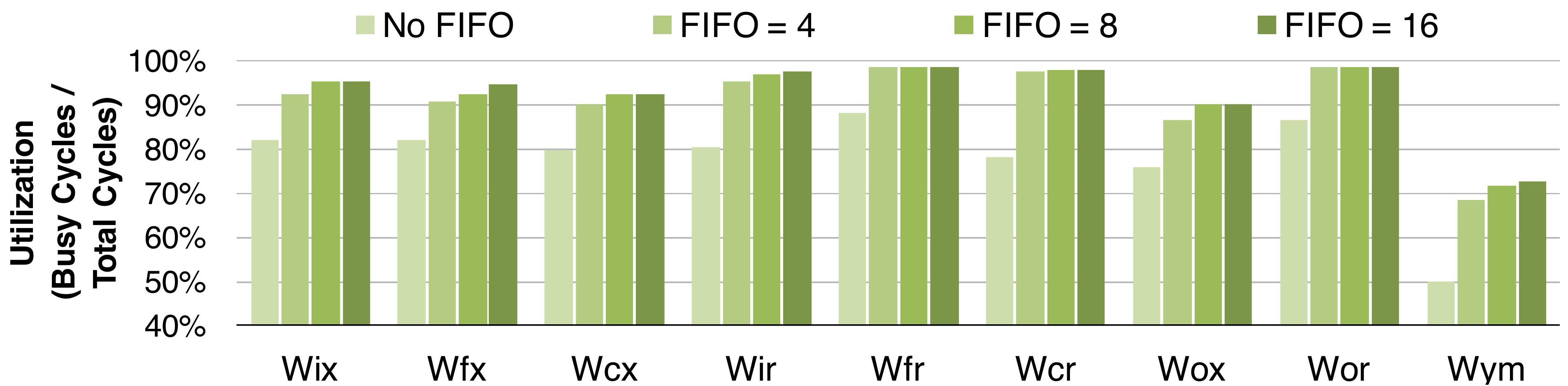}}
\vspace{-5pt}
\caption{FIFO improves load balancing and decreases latency. The ALU utilization is more than 90\% when FIFO depth is 8 for load balancing. }
\label{fig:fifo}
\vspace{-10pt}
\end{figure}

\textbf{FIFO Depth.} ESE uses FIFO to decouple the PEs and solves the load imbalance problem. Load imbalance here means the number of non-zero weight assigned to every PE is different. The FIFO for each PE reduces the waiting time for PEs with fewer computations. We adjust the cache depth to investigate its effect. The FIFO width is 16-bit, and its depth is set at 1, 4, 8, 16. In Fig.\ref{fig:fifo}, when the FIFO depth is one (no FIFO), the utilization, which is defined as busy cycle divided by total cycles, is low (80\%) due to load imbalance. When the FIFO depth is 4, the utilization is above 90\%. When the FIFO depth is increased to 8 and 16, the utilization increased but has a marginal gain. Thus we choose the FIFO depth to be 8. Note that even when the FIFO depth is 8, the last matrix ($W_{ym}$) still has low utilization. This is because that matrix has very few rows and each PE has few elements, and thus the FIFO cannot fully solve this problem for this matrix.

% ===========================
\begin{table}[t]
\vspace{-10pt}
\caption{ESE Resource Utilization.}
%\centering
%\setlength{\tabcolsep}{3pt}
%\scriptsize
\begin{threeparttable}
\begin{tabular}{|c|c|c|c|c|c|}
\hline
 & LUT & LUTRAM\tnote{1} &FF & BRAM\tnote{1} & DSP     \\ \hline
Avail. & 331,680 & 146,880 & 663,360 & 1,080 & 2,760
\\ \hline
Used       &293,920   &69,939    &453,068     &947        &1,504         \\
\hline
Utili.    &88.6\%  &47.6\%  &68.3\%      &87.7\%      &54.5\%        \\
\hline
\end{tabular}
\label{tab:ResourceUtilization}
\begin{tablenotes}
    \item[1] LUTRAM is 64b each, BRAM is 36Kb each.
\end{tablenotes}
\end{threeparttable}
\vspace{-5pt}
\end{table}
% ===========================

\subsection{Accuracy, Speed and Energy Efficiency}

\begin{figure}[t!]
\centering
% \vspace{-10pt}
\scalebox{1}[1]{\includegraphics[width=0.5\textwidth]{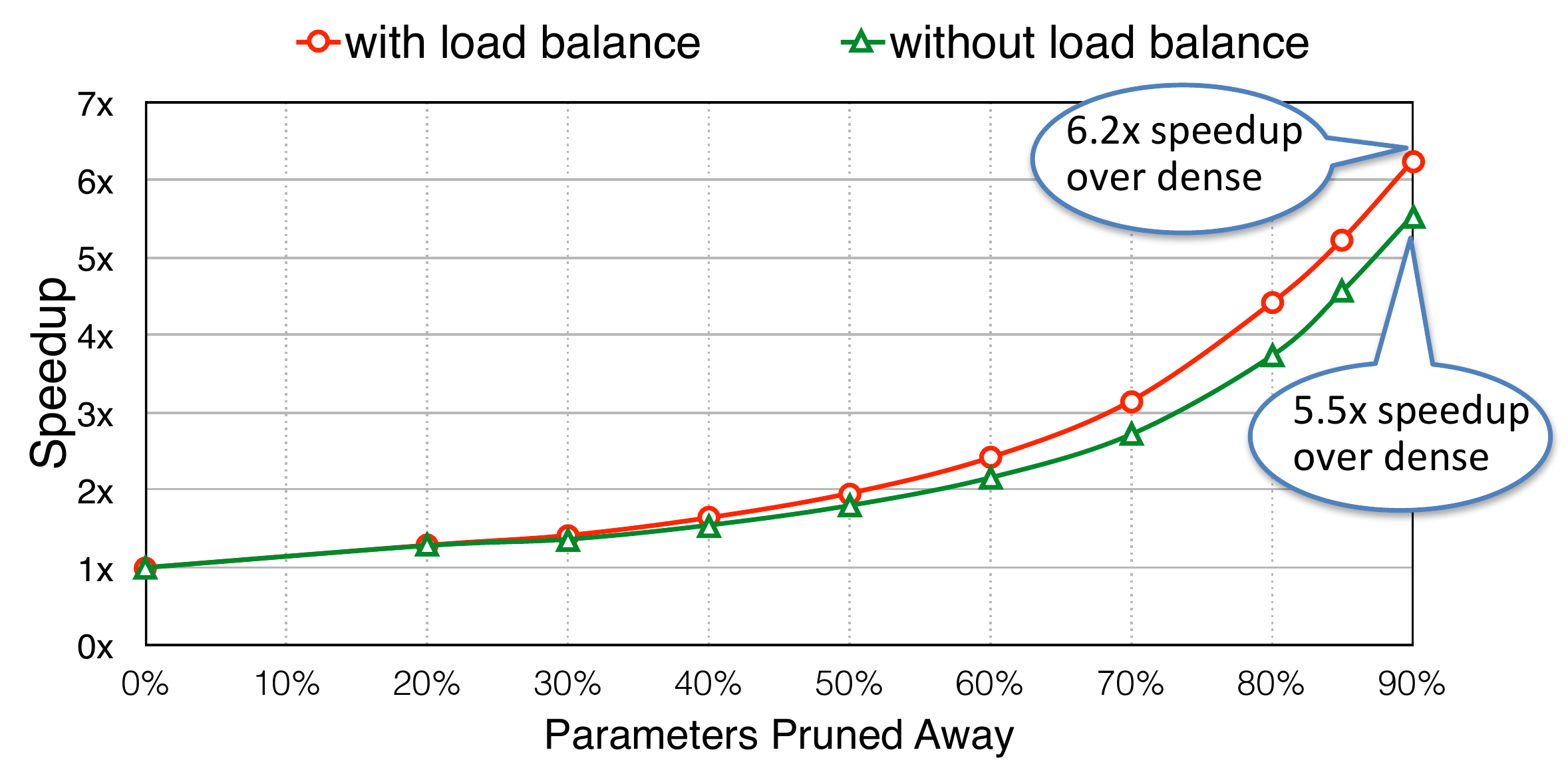}}
% \vspace{-5pt}
\caption{Computation latency decreases as the sparsity increases. Running the sparse model is 4.2$\times$ faster over the dense model, both run on ESE. Load balance aware pruning helps speedup. }
\label{fig:speedup}
% \vspace{-5pt}
\end{figure}

We evaluate the trade-off between accuracy and speedup of ESE in Fig.\ref{fig:speedup}. The speedup increases as more parameters get pruned away. The sparse model which is pruned to 10\% achieved 6.2$\times$ speedup over the dense baseline model. Comparing the red and green lines, we find that load-balance-aware pruning improves the speedup from 5.5$\times$ to 6.2$\times$.

\begin{table}[t]
\vspace{-10pt}
\centering
\caption{Power consumption of different platforms.}
\label{tab:power_comp}
\begin{tabular}{|c|c|c|c|c|c|}
\hline
Platform & CPU & CPU & GPU & GPU & ESE \\
& Dense & Sparse & Dense & Sparse & \\
\hline
Power & 111W & 38W & 202W & 136W & \bf{41W} \\
\hline
\end{tabular}
\end{table}

\begin{figure}[t]
\centering
% \vspace{-5pt}
\scalebox{1}[1]{\includegraphics[width=0.3\textwidth]{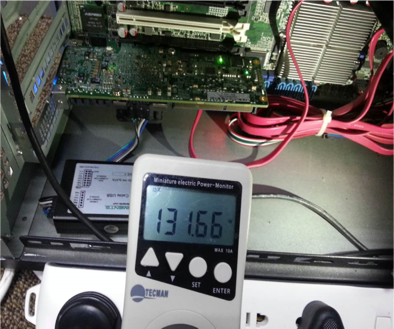}}
% \vspace{-5pt}
\caption{Measured at the socket, the total power consumption of the machine with FPGA fully loaded is 132W. Without FPGA the idle machine consumes 91W. Subtracting the two, ESE consumes 41W.}
\label{fig:power}
\vspace{-10pt}
\end{figure}

We measured power consumption of CPU, GPU and ESE. CPU power is measured by the \texttt{pcm-power} utility. GPU power is measured with \texttt{nvidia-smi} utility. We measure the power consumption of ESE by taking difference with / without the FPGA board installed. ESE takes 41 watts; CPU takes 111 watts (38 watts when using MKLSparse) and GPU takes 202 watts (136 watts when using cuSparse).

\begin{table*}[t]
% \vspace{-25pt}
\centering
\caption{Performance comparison of running LSTM on ESE, CPU and GPU}% (first layer using load balanced-aware pruning with expected compression rate 0.15)}
\label{tab:performance}
%\vspace{-5pt}
\setlength{\tabcolsep}{1pt}
\scriptsize
\begin{threeparttable}
\begin{tabular}{|c|c|c|c|c|c|c|c|c|c|c|c|c|c|}
\hline
{\bf Plat.} & \multicolumn{9}{c|}{\bf ESE on FPGA (ours)} & \multicolumn{2}{c|}{\bf CPU} & \multicolumn{2}{c|}{\bf GPU}
\\ \hline
\multirow{3}{*}{\bf Matrix} & \multirow{2}{*}{\bf Matrix} & \multirow{2}{*}{\bf Sparsity} & {\bf Compres.} 
 & {\bf Theoreti.} & {\bf Real} & {\bf Total} & {\bf Real} & {\bf Equ.} & {\bf Equ.}
 & \multicolumn{2}{c|}{\bf Real Comput.} & \multicolumn{2}{c|}{\bf Real Comput.}
\\
 & \multirow{2}{*}{\bf Size} & \multirow{2}{*}{\bf (\%)\tnote{1}} & {\bf Matrix} 
 & {\bf Comput.} & {\bf Comput.} & {\bf Operat.} & {\bf Perform.} & {\bf Operat.} & {\bf Perform.}
 & \multicolumn{2}{c|}{\bf Time (\si{\bf \micro\second})} & \multicolumn{2}{c|}{\bf Time (\si{\bf \micro\second})}
\\ \cline{11-14}
 & & & {\bf (Bytes)\tnote{2}} 
 & {\bf Time \si{\bf (\micro\second)}} & {\bf Time \si{\bf (\micro\second)}} & {\bf (\si{\bf \giga OP})} & {\bf (\si{\bf \giga OP\per\second})} & {\bf (\si{\bf \giga OP})} & {\bf (\si{\bf \giga OP\per\second})}
 & {\bf Dense} & {\bf Sparse} & {\bf Dense} & {\bf Sparse}
\\ \hline
$W_{ix}$ & 1024$\times$153 & 11.7 & 36608 & 2.9 & \bf{5.36} & 0.0012 & 218.6 & 0.010 & 1870.7 & \multirow{4}{*}{\bf 1518.4\tnote{3}} & \multirow{4}{*}{\bf 670.4}  & \multirow{4}{*}{\bf 34.2} & \multirow{4}{*}{\bf 58.0}
\\ \cline{1-10}
$W_{fx}$ & 1024$\times$153 & 11.7 & 36544 & 2.9 & \bf{5.36} & 0.0012 & 218.2 & 0.010 & 1870.7 & & & &
\\ \cline{1-10}
$W_{cx}$  &  1024$\times$153 & 11.8 & 37120 & 2.9 & \bf{5.36} & 0.0012 & 221.6 & 0.010 & 1870.7 & & & &
\\ \cline{1-10}
$W_{ox}$  & 1024$\times$153 & 11.5 & 35968 & 2.8 & \bf{5.36} & 0.0012 & 214.7 & 0.010 & 1870.7 & & & &
\\ \hline
$W_{ir}$  &  1024$\times$512 & 11.3 & 118720 & 9.3 & \bf{10.31} & 0.0038 & 368.5 & 0.034 & 3254.6 & \multirow{4}{*}{\bf 3225.0\tnote{4}} & \multirow{4}{*}{\bf 2288.0} & \multirow{4}{*}{\bf 81.3} & \multirow{4}{*}{\bf 166.0}
\\ \cline{1-10}
$W_{fr}$  & 1024$\times$512 & 11.5 & 120832 & 9.4 & \bf{10.01} & 0.0039 & 386.3 & 0.034 & 3352.1 & & & &
\\ \cline{1-10}
$W_{cr}$ & 1024$\times$512 & 11.2 & 117760 & 9.2 & \bf{9.89} & 0.0038 & 381.2 & 0.034 & 3394.5 & & & &
\\ \cline{1-10}
$W_{or}$ & 1024$\times$512 & 11.5 & 120256 & 9.4 & \bf{10.04} & 0.0038 & 383.5 & 0.034 & 3343.7 & & & &
\\ \hline
$W_{ym}$ & 512$\times$1024 & 10.0 & 104832 & 8.2 & \bf{15.66} & 0.0034 & 214.2 & 0.034 & 2142.7 & 1273.9 & 611.5 & 124.8 & 63.4
\\ \hline
{\bf Total} & {\bf 3248128} & {\bf 11.2} & {\bf 728640} & {\bf 57.0} & {\bf 82.7} & {\bf 0.0233} & {\bf 282.2} & {\bf 0.208} & {\bf 2515.7} & {\bf \,\,6017.3\,\,} & {\bf 3569.9} & {\bf 240.3} & {\bf 287.4}
\\ \hline
\end{tabular}
\begin{tablenotes}
    \item[1] Pruned with 10\% sparsity, but padding zeros incurred about 1\% more non-zero weights.
    \item[2] Sparse matrix index is included, and weight takes 12 bits, index takes 4 bits => 2 Bytes per weight in total.
    \item[3] Concatenating $W_{ix}$, $W_{fx}$, $W_{cx}$ and $W_{ox}$ into one large matrix $W_{\text{ifoc\_x}}$, whose size is 4096$\times$153.
    \item[4] Concatenating $W_{ir}$, $W_{fr}$, $W_{cr}$ and $W_{or}$ as one large matrix $W_{\text{ifoc\_r}}$, whose size is 4096$\times$512. These matrices don't have dependency and combining matrices can achieve 2$\times$ speedup on GPU due to better utilization.
\end{tablenotes}
\end{threeparttable}
% \vspace{-10pt}
\end{table*}

The performance comparison of LSTM on ESE, CPU, and GPU is shown in Table~\ref{tab:performance}. The CPU implementation used MKL BLAS and MKL SPBLAS for dense/sparse implementation, and the GPU implementation used cuBlas and cuSparse. We optimized the CPU/GPU speed by combining the four matrices of the i, f, o, c gates that have no dependency into one large matrix.  Both mklSparse and cuSparse implementation results in significant lower utilization of peak CPU/GPU performance for the interested matrix size (relatively small) and sparsity (around 10\% non-zeros). We implemented the whole LSTM on ESE. The model was pruned to 10\% non-zeros. There are 11.2\% non-zeros taking padding zeros into account. On ESE, the total throughput is 282 GOPS with the sparse LSTM, which corresponds to 2.52 TOPS on the dense LSTM. Processing the LSTM with 1024 hidden elements, ESE takes 82.7 us, CPU takes 6017.3/3569.9 us (dense/sparse), and GPU takes 240.2/287.4 us (dense/sparse). With batch=32, CPU sparse is faster than dense because CPU is good at serial processing, while GPU sparse is slower than dense because GPU is throughput oriented. With no batching, we observed both CPU and GPU are faster for the sparse LSTM because the saving of memory bandwidth is more salient.

Performance wise, ESE is 43$\times$ faster than CPU 3$\times$ faster than GPU. Considering both performance and power consumption, ESE is 197.0$\times$/40.0$\times$ (dense/sparse) more energy efficient than CPU, and 14.3$\times$/11.5$\times$ (dense/sparse) more energy efficient than GPU. Sparse LSTM makes both CPU and GPU more energy efficient as well, which shows the advantage of our pruning technique.

\section{Related Work}
\textbf{Deep Compression}
Deep Compression~\cite{han2015deep} is a method that can compress convolutional neural network models by 35x-59x without hurting the accuracy. It is comprised of pruning, weight sharing and Huffman coding. However, the compression rate targets CNN and image recognition. In this work we target LSTM and speech recognition. The method also differs from the previously proposed `Deep Compression' in that we proposed load-balance-aware pruning. During pruning, we enforce each row has the same amount of weight to enforce hardware load balancing. During quantization, we use linear quantization instead of non-linear quantization, which is simpler but has smaller compression ratio. We also eliminate the Huffman Coding step which introduces extra decoding overhead but marginal gain. 

\textbf{CNN Accelerators} 
Many custom accelerators have been proposed for CNNs. DianNao~\cite{diannao} implements an array of multiply-add units to map large DNN onto its core architecture. Due to limited SRAM resource, the off-chip DRAM traffic dominates the energy consumption. DaDianNao~\cite{dadiannao} and ShiDianNao~\cite{shidiannao} eliminate the DRAM access by having all weights on-chip (eDRAM or SRAM). However, these DianNao-series architectures are  proposed to accelerate CNNs, and the weights are uncompressed and stored in the dense format. In this work, we target LSTM neural network and speech recognition, and data compression is also supported in our ESE architecture. Our work in this paper also distinguishes itself from Angel-Eye architecture, which also has the compression, compilation and acceleration, but it is accelerating CNNs, not LSTMs~\cite{angeleye,fpga2016cnn}.

\textbf{EIE Accelerator} 
The EIE architecture proposed by Han et al.~\cite{han2016eie} can performs inference on compressed network model and accelerates the resulting sparse matrix-vector multiplication
with weight sharing. With only 600mW power consumption, EIE can achieve 102 GOPS processing power on a compressed network corresponding to 3 TOPS/s on an uncompressed network, which is 24000$\times$ and 3400$\times$ more energy efficient than a CPU and GPU respectively.  EIE is a general building block for deep neural network, not specially designed for LSTM and speech recognition; ESE in this paper targets LSTM. ESE has different design constrains on FPGA while EIE is for ASIC, which leads different design considerations. Besides, EIE uses codebook-based quantization, which has better compression ratio; ESE uses linear quantization, which is easier to implement.

\textbf{Sparse Matrix-Vector Multiplication Accelerators}
To pursue a better computational efficiency on machine learning and deep learning, several recent works focus on using FPGA as an accelerator for Sparse Matrix-Vector Multiplication (SpMV). Zhuo et al.~\cite{Zhuo2005_fpga} proposed an FPGA-based design on Virtex-II Pro for SpMV. Their design outperforms general-purpose processors. Fowers et al.~\cite{Fowers2014_fccm} proposed a novel sparse matrix encoding and an FPGA-optimized architecture for SPMV. With lower bandwidth, it achieves 2.6$\times$ and 2.3$\times$ higher power efficiency over CPU and GPU respectively while having lower performance due to lower memory bandwidth. Dorrance et al.~\cite{Dorrance2014_fpga} proposed a scalable SMVM kernel on Virtex-5 FPGA. It outperforms CPU and GPU counterparts with >300$\times$ computational efficiency and has 38-50$\times$ improvement in energy efficiency.
For compressed deep networks, previously proposed SpMV accelerators can only exploit the static weight sparsity. In this paper, we use the relative indexed compressed  sparse column (CSC) format for data storing, and we develop a scheduler which can map a complicate LSTM network on ESE accelerator.

\textbf{GRU on FPGA} 
Nurvitadhi et al presented a hardware accelerator for Gated Recurrent Network (GRU) on Stratix V and Arria 10 FPGAs~\cite{intel2016_fpl}. This work shows that FPGA can provide superior performance/Watt over CPU and GPU. In our work, we present a FPGA accelerator for LSTM network. It also demonstrates a higher efficiency FPGA comparing with CPU
and GPU. Different from theirs, ESE is especially designed for accelerating sparse LSTM model.

\textbf{LSTM on FPGA}
In order to explore the parallelism for RNN/LSTM, Chang presented a hardware implementation of LSTM network on Zynq 7020 FPGA from Xilinx with 2 layers and 128 hidden units in hardware~\cite{lstm_fpga}. The implementation is 21 times faster than the ARM Cortex-A9 CPU embedded on the Zynq 7020 FPGA. Lee accelerated RNNs using massively parallel processing
elements (PEs) for low latency and high throughput on FPGA \cite{lee2016fpga}. These implementations did not support sparse LSTM network, while our ESE can achieve more speed up by supporting sparse LSTM.

\section{Conclusion}

In this paper, we present Efficient Speech Recognition Engine (ESE) that works directly on compressed sparse LSTM model. ESE is optimized across the algorithm-hardware boundary: we first propose a method to compress the LSTM model by 20$\times$ without sacrificing the prediction accuracy, which greatly saves the memory bandwidth of FPGA implementation. Then we design a scheduler that can map the complex LSTM operations on FPGA and achieve parallelism. Finally we propose a hardware architecture that efficiently deals with the irregularity caused by compression. Working directly on the compressed model enables ESE to achieve 282 GOPS (equivalent to 2.52 TOPS for dense LSTM) on Xilinx XCKU060 FPGA board. ESE outperforms Core i7 CPU and Pascal Titan X GPU by factors of  $43\times$ and $3\times$ on speed, and it is 40$\times$ and 11.5$\times$ more energy efficient than the CPU and GPU respectively.

\section{Acknowledgment}
This work was supported by National Natural Science Foundation of China (No.61373026, 61622403, 61261160501).

We would like to thank Wei Chen, Zhongliang Liu, Guanzhe Huang, Yong Liu, Yanfeng Wang, Xiaochuan Wang and other researchers from Sogou for their suggestions and providing real-world speech data for model compression performance test.
\bibliographystyle{abbrv}
\bibliography{ref}

\end{document}